\documentclass[letterpaper, 10 pt, conference]{ieeeconf}  

\IEEEoverridecommandlockouts                              

\usepackage{style}

\title{\LARGE \bf
Terrain Consistent Reference-Guided RL for Humanoid Navigation Autonomy
}

\author{William D. Compton$^{1}$, Zachary Olkin$^{1}$, and Aaron D. Ames$^{1}$
\thanks{*This research is supported by Technology Innovation Institute (TII).}
\thanks{$^{1}$Authors are with the Department of Computing and Mathematical Sciences, California Institute of Technology,
Pasadena, CA 91125.
}
}

\begin{document}
\bstctlcite{IEEEexample:BSTcontrol}

\maketitle
\thispagestyle{empty}
\pagestyle{empty}

\begin{abstract}

We present a method for training reference-guided, perceptive reinforcement learning locomotion policies for humanoid robots in which reference trajectories are modulated in training to be consistent with terrain geometry.
Aiming to deploy our method with standard navigation autonomy infrastructure, we synthesize $SE(2)$-controllable reference trajectories inside the RL training loop, projecting desired footsteps onto valid footholds and adjusting swing-foot and center-of-mass trajectories to match the terrain.
The resulting policy exposes a clean $SE(2)$ velocity interface compatible with standard navigation planners. 
In simulation, environmentally-conditioned references significantly improve reference tracking performance compared to environment agnostic references.
On hardware, we integrate the policy with an MPC + control barrier function planner and demonstrate long-horizon ($>$70m) closed-loop autonomous navigation on the Unitree G1 through outdoor environments containing rough terrain and consecutive flights of stairs, with all sensing and computation onboard.
\end{abstract}

\section{Introduction}
As contemporary locomotion methods for humanoid robots achieve high levels of agility and robustness, the bottleneck increasingly lies in deploying these behaviors as part of a larger navigation stack in complex and unstructured environments.
While embedding motion priors from human data or trajectory optimization into RL training has led to state-of-the-art performance and robustness for humanoid locomotion, these methods struggle to address two main problems: interfacing with planner-issued commands and conditioning on the terrain geometry such as valid footholds. 
Our method tackles both of these shortcomings, training a reference-shaped, perceptive locomotion policy designed to integrate with navigation architectures by leveraging reduced order models to synthesize environmentally consistent, $SE(2)$ controllable reference signals in the RL training loop.

Controllability and environmental conditioning of the locomotion policy are critical components of any general humanoid navigation stack. 
Environmental factors such as stairs, constrained footholds, and tight spaces necessitate tight coupling with perception.
Jointly addressing low-level stabilization and long horizon, non-convex and terrain dependent problem of navigation exceeds modern resource budgets on compute, data, and observability.
These limitations necessitate the division between navigation planning, typically providing waypoints or paths, from locomotion control, tasked with realizing the navigation plan.
Therefore, the locomotion policy must expose a planner-compatible, perception-aware interface.

In the context of humanoid robots, training agile locomotion often leverages human data or trajectory optimization to guide the RL.
By training the policy to follow trajectories obtained from human data or trajectory optimization, emergent undesirable behaviors are substantially reduced.
This approach can leverage the significant capabilities and inherent robustness of reference motions to generate more agile, dynamic, and robust motions than possible with other reference-free methods.

\begin{figure}
    \centering
    \includegraphics[width=1.0\linewidth]{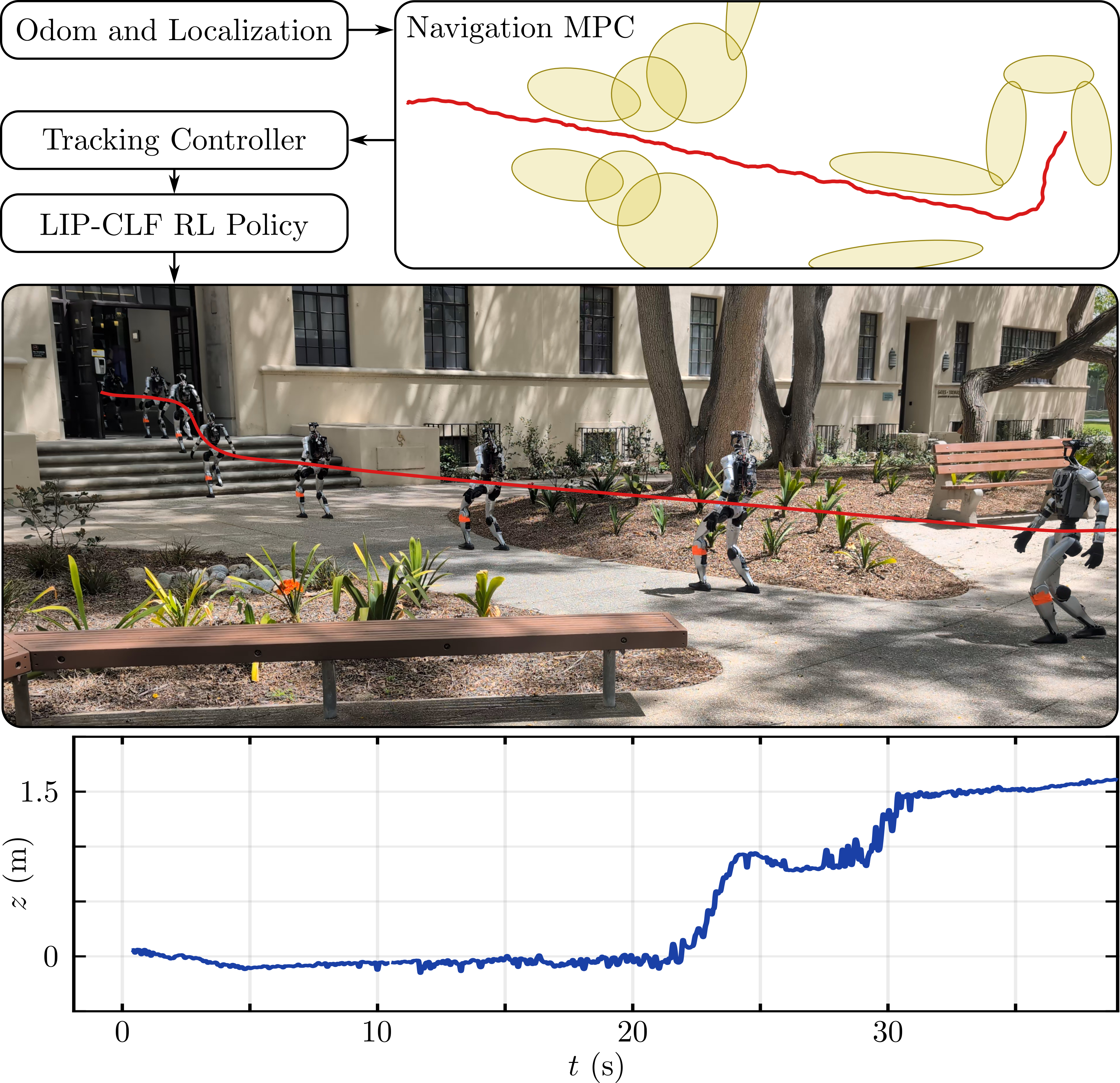}
    \caption{LIP-CLF RL trains a unstructured terrain locomotion policy using a single stage, MLP based training pipeline and a single depth camera. The policy is $SE(2)$ controllable, enabling integration with standard navigation planning methods. In the pictured experiment, the Unitree G1 robot autonomously navigates into a building, climbing two flights of stairs. Video available at \url{https://youtu.be/rTWgR_VxL5Y}.}
    \label{fig:hero}
    \vspace{-5mm}
\end{figure}

\begin{figure*}
    \centering
    \includegraphics[width=\linewidth]{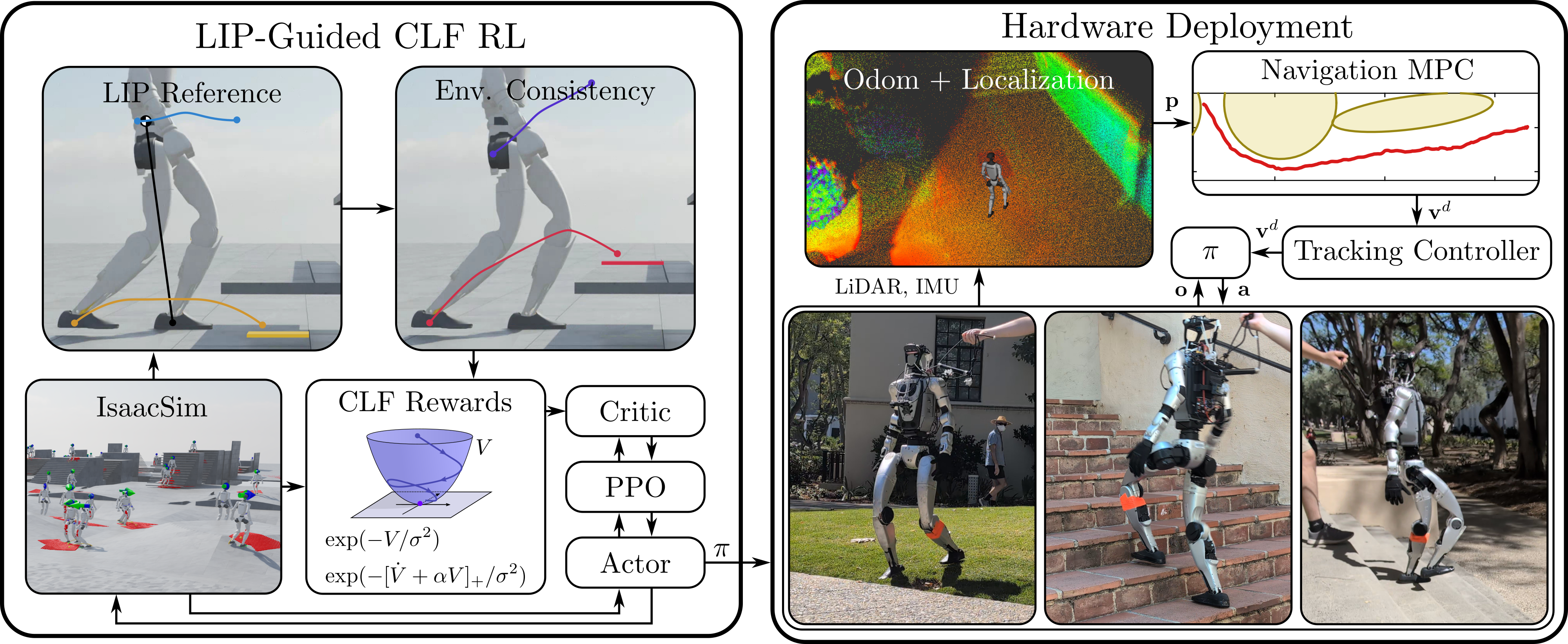}
    \caption{System Architecture Diagram. References are generated in the CLF-RL training loop, and modified to be consistent with the environment. When deployed on hardware, lidar is used for odometry and localization in a map, and a ZED Mini mounted on the robot's chest obtains depth images. MPC plans trajectories for the robot; a tracking controller sends velocity commands to the RL policy to realize the plan.}
    \label{fig:architecture}
    \vspace{-5mm}
\end{figure*}

While reference-guided RL makes significant progress on the agile locomotion front, it is often not conducive with integration into navigation planning architectures. 
Many approaches require reference trajectories as inputs, often live motion-capturing and imitating human motions.
Either synthesizing environmentally consistent whole-body references online or sequencing discrete capabilities to accomplish general locomotion tasks are prohibitively challenging planning problems on their own.
This lack of an interface with planning methods limits the applicability of these methods in standard navigation architectures.

Additionally, conditioning reference-guided RL pipelines on environmental factors, such as terrain geometry, remains a difficult problem. 
Standard motion-capture pipelines do not collect terrain information, and even if references are generated with terrain geometry, it is nontrivial to map these references on to new terrains not in the reference library.

Our proposed method is designed to solve the problem of reference-guided perceptive locomotion capable of integrating with existing planning architectures. To accomplish this, we propose a novel method leveraging reduced order models to synthesize environmentally consistent reference signals rapidly in the RL training loop. The policy tracks $SE(2)$ velocity commands, which enables direct integration with standard navigation planning architectures, shown in \cref{fig:architecture}. Our policy is conditioned directly on depth camera data, allowing robust navigation over complex and unstructured terrain, which we demonstrate on the Unitree G1. 

Specifically, we make the following contributions:
\begin{itemize}
\item We introduce an environment-conditioned reference synthesis method based on the Linear Inverted Pendulum (LIP) model that runs inside the RL training loop.
To our knowledge, this is the first reference-guided RL pipeline for humanoids in which references are modulated online to be consistent with terrain geometry. Further, our policy uses only an MLP, no history, and can be trained without distillation, vastly simplifying the learning problem relative to other state of the art methods.
\item We integrate the resulting perceptive locomotion policy into a complete navigation autonomy stack via a classic $SE(2)$ velocity interface, demonstrating outdoor long-horizon ($>$70m) closed-loop autonomous navigation through human-centric environments including stairs, rough terrain, and flat ground.
As shown in \cref{tab:comparison}, prior humanoid navigation autonomy work has relied on blind, model-based locomotion controllers, while existing perceptive RL locomotion policies have not been deployed in autonomy stacks at this scale.
\end{itemize}

\section{Related Work}

\begin{table*}[t]
\centering
\caption{Comparison of related works. ``Method characterization'' columns describe the policy and its training; ``Deployment characterization'' columns describe how the policy is used downstream. ``Env. cond.'' refers to environmental conditioning of the reference.}
\label{tab:comparison}
\renewcommand{\arraystretch}{1.0}
\setlength{\tabcolsep}{4pt}
\small
\begin{tabular}{@{}llcccc cccc@{}}
\toprule
& & \multicolumn{4}{c}{\textbf{Method characterization}} & \multicolumn{4}{c}{\textbf{Deployment characterization}} \\
\cmidrule(lr){3-6} \cmidrule(lr){7-10}
\textbf{Method} & \textbf{Robot} & \textbf{Ref. source} & \textbf{Env. cond.} & \textbf{Perception} & & \textbf{Interface} & \textbf{Auto. Nav.} & \textbf{Long horiz.} & \textbf{Terrain}$^\dagger$ \\
\midrule
\multicolumn{10}{@{}l}{\textit{Perceptive locomotion}} \\
DTC \cite{jenelten2024dtc}                          & ANYmal & TO (in loop)      & \checkmark & Heightmap   & & Path/footstep & \xmark & \xmark & Urban \\
BeamDojo \cite{wangBeamDojoLearningAgile2025}       & G1     & ---               & ---        & Heightmap   & & SE(2) vel.    & \xmark & \xmark & Custom     \\
Gallant \cite{benGallantVoxelGridbased2025}         & G1     & ---               & ---        & Voxel grid  & & Goal pose     & \xmark & \xmark & Custom     \\
PIM \cite{long2025learning}                         & Various & ---              & ---        & Heightmap   & & SE(2) vel.    & \xmark & \xmark & Urban \\
AME \cite{he2025attention}                          & ANYmal & ---               & ---        & Heightmap   & & SE(2) vel.    & \xmark & \xmark & Custom \\
RPL \cite{zhang2026rpl}                             & G1     & ---               & ---        & Depth       & & SE(2) vel.    & \xmark & \xmark & Urban \\
PHP \cite{wu2026perceptive}                         & G1     & Mocap             & \xmark     & Depth       & & SE(2) vel.    & \xmark & \xmark & Parkour     \\
Su \emph{et al.} \cite{su2025lipm}                  & Biped  & LIPM$^\ddagger$ (in loop) & \xmark     & Depth       & & SE(2) vel.    & \xmark & \xmark & Urban \\
\midrule
\multicolumn{10}{@{}l}{\textit{Reference-tracking methods}} \\
BeyondMimic \cite{liao2025beyondmimic}              & G1       & Mocap           & \xmark     & ---         & & Full traj.    & \checkmark & \xmark & Flat     \\
ZEST \cite{sleiman2026zest}                         & Atlas/G1    & Mocap        & \xmark     & ---         & & Full traj.    & \xmark & \xmark & Parkour     \\
ASAP \cite{lee2024asap}                             & G1       & Video           & \xmark     & ---         & & Full traj.    & \xmark & \xmark & Flat     \\
VideoMimic \cite{allshire2025visual}                & G1       & Video           & \xmark     & Heightmap   & & Full traj.    & \xmark & \xmark & Urban     \\
Opt2Skill \cite{liu2025opt2skill}                   & Digit    & TO              & \xmark     & ---         & & Task space    & \xmark & \xmark & Flat     \\
CLF-RL \cite{li2026clf, olkin2026chasing}            & G1       & TO + Mocap      & \xmark     & ---         & & SE(2) vel.    & \checkmark & \xmark & Flat     \\
\midrule
\multicolumn{10}{@{}l}{\textit{Humanoid navigation autonomy}} \\
Huang \emph{et al.} \cite{huang2023efficient}       & Cassie$^*$   & ---             & --- & Heightmap & & Path     & \checkmark & \checkmark     & Rough     \\
StateNav \cite{yoon2025state}             & Digit$^*$   & ---             & --- & Heightmap & & Path     & \checkmark & \xmark     & Rough \\
Xiong \emph{et al.} \cite{xiongGlobalPositionControl2021} & Cassie$^*$ & ---         & --- & None & & Goal pose & \checkmark & \xmark     & Flat     \\
NavILA \cite{cheng2024navila} & G1/Booster & ---         & --- & Heightmap & & SE(2) vel. & \checkmark & \xmark     & Flat     \\
\midrule
\textbf{Ours}                                       & \textbf{G1} & \textbf{LIP (in loop)} & \textbf{\checkmark} & \textbf{Depth} & & \textbf{SE(2) vel.} & \textbf{\checkmark} & \textbf{\checkmark} & \textbf{Urban} \\
\bottomrule
\multicolumn{10}{@{}p{\dimexpr\linewidth-2\tabcolsep}@{}}{\footnotesize $^{\dagger}$Terrains deployed on. Rough indicates non-flat terrain. Custom indicates terrains built in lab. Parkour indicates use of non-foot terrain contact. Urban indicates deployment in human environments. $^{\ddagger}$Reference is CoM position only. $^*$Locomotion controller model-based, operating on ALIP or LIP models.}
\end{tabular}
\vspace{-5mm}
\end{table*}

Humanoid locomotion has a relatively long history, ranging from model-based works leveraging structured perception to reinforcement learning methods acting on raw sensor data.
We organize related work into three categories, summarized in \cref{tab:comparison}. First, perceptive locomotion methods that achieve robust hardware deployment, without reference trajectories, which have yet to demonstrate integration with navigation autonomy stacks. 
Second, reference-tracking methods which lack environmental conditioning and typically require full-DoF reference trajectories at inference time, and are thus not compatible with standard navigation planning techniques.
Third, autonomous humanoid navigation systems, which close the autonomy gap, but lack strong perceptive locomotion, and cannot traverse common urban and human-centric terrains such as stairs. 
As we discuss below, no prior work known to the authors lies at this intersection: perceptive, reference guided RL-based locomotion deployed in a closed-loop navigation autonomy stack in unstructured human-centric environments. 

\subsection{Perceptive RL Locomotion}
Locomotion over rough terrains such as stairs or discrete footholds represents a key advantage for legged robots over other locomotion modalities. To handle rough terrain, state-of-the-art approaches leverage environment sensing, typically through either depth cameras or LiDAR units. Several methods construct environment representations such as height-maps or occupancy grids, which can be queried as policy inputs \cite{long2025learning,he2025attention,wangBeamDojoLearningAgile2025, benGallantVoxelGridbased2025}. 
Alternatively, sensor data can be used as raw inputs to the policy; in this case, recursive or history-based policies are typically used to deal with limitations of raw sensor data including occlusion or partial observability \cite{zhuang2024humanoid, su2025lipm, zhang2026rpl}.
While such methods have achieved incredible robustness and agility in unstructured environments, it is worth noting that, as opposed to reference tracking pipelines, reference-free pipelines often give rise to undesirable behaviors such as period-two gaits, undesired environmental contacts such as stair edges or risers, or un-anthropomorphic behaviors. 
While some work exists on leveraging references to guide RL in unstructured environments for quadrupeds \cite{jenelten2024dtc} or very recently on humanoids \cite{wu2026perceptive}, it is a largely undeveloped field for humanoid robots due to the complexity of generating environmentally consistent references. As far as the authors are aware, no work exists where humanoid locomotion is trained using reference-guided RL where the references are modulated to be environmentally consistent during training.

\subsection{Reference-Guided RL for Humanoid Locomotion}
Rather than learning complex motion from scratch, many works leverage reference trajectories to guide the reinforcement learning. 
These are typically seeded with human data, \cite{liao2025beyondmimic, sleiman2026zest, allshire2025visual, lee2024asap}, or trajectory optimization \cite{li2026clf, liu2025opt2skill, olkin2026chasing, yu2022dynamic}.
Most methods require a full body trajectory during execution as an input to track, potentially balanced with goal conditioning to accomplish behavioral goals.
Requiring reference trajectories for deployment presents a significant bottleneck when deploying the controller in an autonomy stack, especially when interactions with the environment are nontrivial. 
Some preliminary work using diffusion models to generate reference trajectories online has been investigated \cite{liao2025beyondmimic}, but has yet to demonstrate navigation beyond short indoor trajectories avoiding a single obstacle on flat ground.
In general, few reference-tracking pipelines are conditioned on environmental perception, and deployment in autonomous navigation is underdeveloped. 

The use of reduced order models to shape rewards has also been investigated, although synthesizing full-body reference trajectories has not been studied. The LIP model was used to plan footsteps \cite{lee2024integrating}, but hardware deployment was blind on flat ground. Perceptive techniques have been investigated on a simple 6-DoF point foot biped \cite{su2025lipm}, where the LIP model is used to incentivize the CoM to stay within stable regions. 

\subsection{Humanoid Navigation Autonomy}
Despite great strides being made in quadruped autonomy \cite{agha2022nebula, dixit2024step}, and humanoid locomotion, relatively few works exist on humanoid navigation autonomy in unstructured environments. The use of model predictive control for humanoid path planning is well studied \cite{xiongGlobalPositionControl2021}, and has been combined with traversability analysis for handling rough terrain \cite{lin2021long, huang2023efficient, yoon2025state}. However, in the context of terrain-aware navigation, nearly all works utilize blind locomotion policies, and handle roughness in the terrain through robustness rather than perception. Modern perception-based locomotion approaches have yet to be deployed in autonomous navigation architectures. The authors hypothesize that non-standard interfaces with higher level planning layers (training only to reach goal positions \cite{benGallantVoxelGridbased2025}, requiring full degree of freedom reference trajectories at runtime \cite{liao2025beyondmimic}, etc.) make this integration into autonomous navigation difficult. Our method interfaces with $SE(2)$ velocity commands, integrating with standard autonomous navigation pipelines. Our deployment on long-horizon navigation tasks through human-centric environments containing both rough terrain and stairs is novel and highlights the utility of our training methodology.

\section{Environmentally Consistent Reference Generation}
This section details the use of the Linear Inverted Pendulum (LIP) model to generate environmentally consistent reference trajectories for the feet, center of mass, and arms.
Humanoid locomotion is inherently underactuated, as the center of mass cannot be arbitrarily controlled due to contact and input constraints.
The LIP model is a standard in humanoid locomotion which models the robot as a point mass vaulting over its stance foot, as shown in \cref{fig:ref_mod}.

\subsection{Gait Synthesis via Linear Inverted Pendulum}
The LIP model is described by a state $\b x_{l} :=[p \, v]^\top$, with $p$ the position of the center of mass with respect to the stance foot, and $v$ its velocity.
Assuming the mass remains at a constant height $z_0$ above the ground, the LIP evolves according to dynamics \cite{xiong3DUnderactuatedBipedal2021}, with $\lambda=\sqrt{g / z_0}$
\begin{equation}
    \dot{\b x}_{l} = \begin{bmatrix}
        0 & 1 \\ 
        \lambda^2 & 0
    \end{bmatrix} \b x_l = \b A \b x_l
\end{equation}
Finally, the LIP model is assumed to step at a fixed period $T = 0.4s$. Upon a step, the position state $p$ is updated to be relative to the new stance foot. Given a step of length $u$, the step-to-step dynamics \cite{xiong3DUnderactuatedBipedal2021} are given by integrating the continuous time dynamics and applying the position reset:
\begin{equation}
    \b x_{l}^+ = \begin{bmatrix} p\cosh(\lambda T) + v\sinh(\lambda T)/\lambda \\ p\lambda \sinh(\lambda T) + v\cosh(\lambda T) \end{bmatrix} - \begin{bmatrix}
        u \\ 0
    \end{bmatrix} = \b A_d \b x_{l} + \b B_d u
\end{equation}
We treat the sagittal and lateral planes of the robot as decoupled LIP models. During steady-state walking, the sagittal plane will evolve according to a period one gait, while the lateral plane is period two.

Next, consider a desired $SE(2)$ velocity command, $\b v^d = [v_{x}^d, \, v_{y}^d, \, \omega_{z}^d]^\top$, where the subscript $x$ corresponds to the sagittal plane, and $y$ the lateral plane. 
The linear velocities correspond directly to desired step lengths in the sagittal and lateral planes, $\b u^d = [u^d_{x}, \, u^d_{y}]^\top$:
\begin{align*}
    u^d_{x} &= v^d_{x}T \\
    u^d_{y,1} &= v^d_{y}T+w\\ u^d_{y,2} &= v^d_{y}T-w
\end{align*}
where $u_{y,1}^d$ and $u_{y,2}^d$ are the first and second steps of the period two lateral gait, and $w > 0$ is the nominal lateral step width.
Given these desired step lengths, the post-impact state of the center of mass can be solved for as the fixed point of the step-to-step dynamics. First, in the sagittal plane:
\begin{align}
    \b x^d_x &= \b A_d \b x^d_x + \b B_d u^d_{x} \\
    \b x^d_x &= (\b I - \b A_d)^{-1} \b B_d u^d_{x}
\end{align}
And second, solving for both steps in the lateral plane:
\begin{align}
    \b x^d_{y,1} &= \b A_d (\b A_d \b x^d_{y,1} + \b B_d u^d_{y, 1}) + \b B_d u_{y,2}^d \\
    \b x^d_{y,1} &= (\b I-\b A_d^2)^{-1}(\b A_d \b B_d u^d_{y, 1} + \b B_d u_{y,2}^d) \\
    \b x^d_{y,2} &= \b A_d \b x^d_{y,1} + \b B_d u_{y,1}^d
\end{align}
Where $\b x^d_x$ is the desired post-impact sagittal position and velocity of the CoM, and $\b x^d_{y,1}$, $\b x^d_{y,2}$ are the desired post-impact lateral position and velocity, for the first and second step respectively. The position of the center of mass at time $t \in [0, T)$ is then recovered from the continuous time dynamics:
\begin{equation}
    \b x^d_{i,j}(t) = \begin{bmatrix} \cosh(\lambda t) & \sinh(\lambda t)/\lambda \\ \lambda \sinh(\lambda t) & \cosh(\lambda t) \end{bmatrix} \b x^d_{i,j}
\end{equation}
where $i$ indexes the sagittal and lateral planes, and $j$ indexes the first and second steps for the period two lateral gait.
Finally, the yaw of the CoM is computed to ensure zero yaw relative to the stance foot during mid-stance:
\begin{equation}
    \psi(t) = \omega_z^d\left(t - \frac{T}{2}\right)
\end{equation}
Given the center of mass trajectory and desired footstep locations, the rest of the humanoid reference trajectory is established heuristically.
The roll and pitch of the root are fixed to zero, as are the arm and waist joints. 
The swing foot trajectory is a bezier curve connecting the initial position of the swing foot, $\b x_{sw,0}$ to the desired step location $\b u^d$:
\begin{equation}
    \b x_{sw}(t) = \b u^d b(t) + (1 - b(t))\b x_{sw,0} \label{eqn:swingfoot}
\end{equation}
with $b(t)$ the unique third order polynomial satisfying
\begin{equation}
    b(0) = 0 \quad \dot{b}(0) = 0 \quad b(T) = 1 \quad \dot{b}(T) = 0
\end{equation}
The $z$ height of the swing foot follows an identical bezier curve up and down, with $\bar{z}_{sw}$ the maximum swing height:
\begin{equation}
    z_{sw}(t) = \begin{cases}
        \bar{z}_{sw} b(2t) & t \in [0, T/2) \\
        \bar{z}_{sw} b(2(T-t)) & t \in [T/2, T)
    \end{cases}
\end{equation}
The roll and pitch of the swing foot are held at zero, and the yaw of the swing foot follows a similar bezier curve, such that it will align with the CoM at the next mid-stance:
\begin{equation}
    \psi_{sw}(t) = \omega_z^d T b(t) + (1 - b(t))\psi_{sw,0}
\end{equation}
The z-height of the center of mass ($z_d(t)$) and the arms ($\theta_d^i(t)$ for $i \in \{l, r\}$) are given sagittal velocity dependent sinusoids:
\begin{align}
    z^d(t) &= z_0 + z_a\frac{v_{x}^d}{\bar{v}_{x}^d} \left( \sin{\left(\frac{2\pi}{T}t + \pi\right)} - 1\right)\\
    \theta^d_i(t) &= \theta_a\frac{v_{x}^d}{\bar{v}_{x}^d} \sin{\left(\frac{2\pi}{T}t + \theta_0^i\right)}
\end{align}
where $z_a, \theta_a$ are the amplitudes at the maximum forward speed, $\bar{v}_{x}^d$, and $\theta_0^i$ are phase offsets such that the arms swing in opposite phase to the legs.

This construction specifies a complete $SE(2)$ controllable reference trajectory for the humanoid robot while walking on flat ground. This construction is not novel, and follows closely from previous work \cite{xiong3DUnderactuatedBipedal2021}, with the addition of heuristic references for the arms and vertical component of the center of mass.

\subsection{Adjustment for Terrain Consistency}
\begin{figure}
    \centering
    \includegraphics[width=1.0\linewidth]{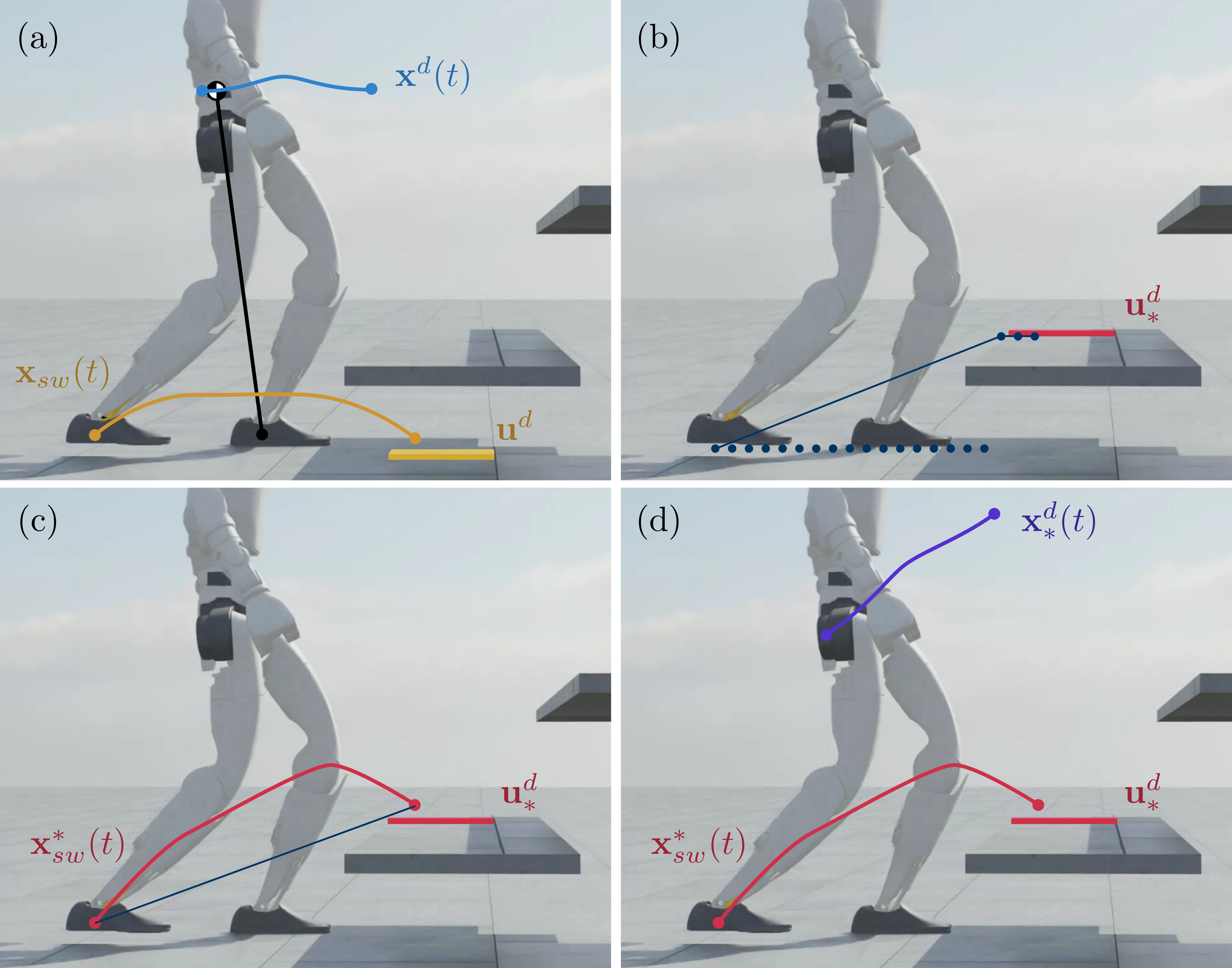}
    \caption{Modification of the LIP reference for terrain consistency. (a) The nominal reference, composed of footstep $\b u^d$, swing foot trajectory $\b x_{sw}(t)$, and pelvis reference $\b x^d(t)$. (b) Projected footstep $\b u^d_*$ and upper convex hull of terrain. (c) Modified swing trajectory $\b s_{sw}^*(t)$ and vertical step deviation line. (d) Modified pelvis trajectory $\b x^d_*(t)$, terrain consistent references.}
    \label{fig:ref_mod}
    \vspace{-5mm}
\end{figure}
The primary contribution of this work is the rapid synthesis of an environmentally consistent reference in the training loop.
To accomplish this, we adjust the CoM and swing foot trajectories to be consistent with the environment over the next step, as pictured in \cref{fig:ref_mod}.
First, we project the desired footstep location onto a valid foothold.
\begin{equation}
    \b u_{*}^d = G^{-1}\left(\text{proj}\left(G(\b u^d, \b x_{st}^W)\right), \b x_{st}^W\right)
\end{equation}
Where $G(a, b)$ takes the $x,y$ position $a$ expressed in the frame of $b$ and expresses it in the world frame, and $\text{proj}(u)$ projects a desired footstep location onto the nearest valid footstep location. $\b x_{st}^W$ is the location of stance foot in the world frame.
To facilitate rapid GPU-parallelized projection, polygons denoting valid footholds are constructed with the terrain.
The terrain is divided into a grid, and each grid cell stores an array of possible polygons to project to from within its grid cell. 
To query the projection operation, the desired foothold is projected onto every polygon in its corresponding grid cell, and the shortest projection is taken.
The roll and pitch ($r_d, p_d$) of the terrain at the projection point are noted. 

The $x, y, \theta, \phi, \psi$ of the swing foot trajectory are all modified to connect the initial foot pose to the desired pose (including the roll and pitch of the terrain at the step location), using the same bezier construction as in \eqref{eqn:swingfoot}.
With the $x,y$ trajectory of the swing foot fixed, the height of the terrain is scanned along the trajectory.
The $z$ swing height is modified by adding the height of the upper convex hull of the terrain points on the swing trajectory. 
Finally, the height of the line joining the current stance foot to the projected step location is added to the $z$ height of the CoM trajectory. 
These modifications to enforce terrain consistency are visualized in \cref{fig:ref_mod}. Importantly, these modifications are general, working for a variety of terrains including stairs, slopes, and discrete footholds, allowing the method to be adapted to any terrain.

\section{Locomotion and Navigation Methodology} \label{sec:method}

We outline a single-stage, MLP-based locomotion training setup which maps the current depth image and proprioception directly to actions. 

\subsection{Reward Design via CLF-RL}
We build on previous work leveraging Control Lyapunov Functions for Reward Design \cite{li2026clf, olkin2026chasing}. Given desired trajectories for the pelvis, swing foot, and joints, we design quadratic Lyapunov functions $V_i(x)$ for each of the tracking objectives. For each tracking objective, we then provide two reward terms, 
\begin{align*}
    r_{i} &= w_{1,i} r_{V_i} + w_{2,i} r_{\dot{V}_i} \\
    r_{V_i} &= \text{exp}(-V_i(x) / \sigma_{1,i}^2) \\
    r_{\dot{V}_i} &= \text{exp}(-[\dot{V}_i(x) + \alpha_i V_i(x)]_+ / \sigma_{2,i}^2)
\end{align*}
Where $w_{j,i} > 0$ are reward weights and $\sigma_{j,i} > 0$ are scaling parameters. 
The first reward encourages tracking the reference signal closely, while the second condition incentivizes satisfying the Lyapunov decrescent condition.
This effectively densifies the reward, allowing the system to achieve some reward even when tracking is poor, provided the system is moving towards the reference as quantified by the Lyapunov function. 

\subsection{Depth Camera Simulation}

The robot is equipped with a single ZED Mini depth camera mounted on the torso, facing downwards. At deployment, the ZED SDK \cite{stereolabs_zed_sdk} runs depth reconstruction on the 1080p image at 35Hz using the 'NEURAL' setting onboard the Jetson Orin integrated with the Unitree G1 robot. The image is then down-sampled via nan-aware averaging to a resolution of 26x30. 
To simulate this sensor in IsaacLab, a 26x30 raycast is performed using a pinhole camera model with intrinsics reported by the ZED Mini camera. Following \cite{benGallantVoxelGridbased2025}, self-scans of the robot geometry (dynamic meshes) are handled by transforming the rays into each mesh's local frame before raycasting:
\begin{equation*}
    \text{raycast}(TM, \b p, \b d) = \text{raycast}(M, T^{-1}\b p, R^{-1} \b d).
\end{equation*}
The goal is to cast a ray starting from position $\b p$, in direction $\b d$, onto a mesh $M$ which has been transformed by homogeneous transform $T$ with rotational component $R$ relative to its mesh frame. We instead compute the raycast by transforming the ray origin into the mesh local frame via $T^{-1}\b p$, and rotating the cast directions into the mesh frame, $R^{-1} \b d$, then casting against a static mesh in its own frame. The result of the raycast is reported in Z-depth to match the depth image computation. 

\subsection{Asymmetric Actor-Critic Implementation}

As the RL pipeline is centered around reward tracking, the reward signals are quite dense, and the training pipeline can be considerably simplified relative to existing humanoid perceptive locomotion pipelines. 
We adopt a single-stage asymmetric actor-critic framework with MLP networks, where the actor is given noised proprioception and depth measurements, as well as noise-free velocity commands and phasing variable. Our depth noise model is simple compared to most (the image is aggressively down-sampled, which smooths edge artifacts and other standard depth image artifacts); we first sample a random bias plane which is added to the entire image, and then add uniform random noise.  The critic is given noise-free observations and additional privileged information including desired reference trajectories and actual outputs of the system, desired and actual contact states, and a ground truth height scan. 

We train on a curriculum-based, randomized terrain consisting of rough terrain, slopes and stairs, pictured in \cref{fig:architecture}. Velocity commands are randomized, consisting of sample-and-hold commands, and PD control to target positions. To facilitate transfer to hardware, we domain randomize across parameters including actuator gains, friction coefficients, mass distributions, disturbances, and joint offsets. 

\subsection{Model Predictive Control for Navigation}

The output of our RL pipeline is a reference-guided perceptive locomotion policy which takes $SE(2)$ velocity commands as input. To solve navigation problems, we pair this locomotion policy with a navigation planner, which takes in an environment representation and a goal pose, and drives the robot along a collision free path to the goal. We use model-predictive control on a single integrator with heading, discretized at $\delta t = 0.4s$, which interfaces naturally with $SE(2)$ velocity commands. 
This model has state $\b z = [x, y, \theta]^\top \in \R^3$ the position and heading, and input $\b v = [v_{\parallel}, v_\perp, \omega]^\top \in \R^3$ the parallel, perpendicular, and angular velocities. The navigation MPC problem is written over horizon $N=25$:
\vspace{-1mm}
\begin{mini!}[1]
{\b v_{(\cdot)}, \b z_{(\cdot)}}{J(\b z_{(\cdot)}, \b v_{(\cdot)}) \label{eq:OCP_objective}}
{\label{eq:OCP}}{}
\addConstraint{\b z_{k+1}}{= f(\b z_{k}, \b v_k) \quad \label{eqn:ocp_dyn}}{k \in [0, N]}
\addConstraint{\b z_0}{= \b \Pi(\b x(t)) \label{eqn:ocp_ic}}
\addConstraint{\b v_k}{\in \mathcal{V} \quad \label{eqn:ocp_input}}{k \in [0, N]}
\addConstraint{ g_i(\b z_k, \b z_{k+1}) }{\geq \delta \quad \label{eqn:ocp_barrier}}{k \in [0, N]}
\end{mini!}

Where the cost $J$ is composed of a geometrically consistent quadratic distance to goal penalty ($c$ handles the cost on heading in a geometrically consistent way, which is locally quadratic) and quadratic input penalty,
\begin{align*}
    J(\b z_{(\cdot)}, \b v_{(\cdot)}) &= c(\b z_N) + \sum_{k=0}^{N-1} \big(c(\b z_k) + \b v_k^\top R \b v_k \big) \\
    c(\b z) &= q_{xy} \left((x - x_g)^2 + (y - y_g)^2\right) + \\ &\mathrel{\phantom{=}} q_{\theta}\left(\sin^2(\theta - \theta_g) + (1 - \cos(\theta - \theta_g))^2 \right)
\end{align*}
where $\b z_g = [x_g, y_g, \theta_g]^\top$ is the goal pose. The single integrator with heading dynamics are enforced in \eqref{eqn:ocp_dyn}, the initial condition constraint \eqref{eqn:ocp_ic} extracts the position and heading via the projection map $\b \Pi: \mathcal{X} \to \R^3$ given by: 
\begin{equation*}
    \b \Pi(\b x) = [x, y, \theta]^\top 
\end{equation*}
Lastly, given functions $h_i: \R^3 \to \R$ encoding distance to each obstacle $i$, we enforce separate discrete-time control barrier constraints \eqref{eqn:ocp_barrier} to perform obstacle avoidance \cite{agrawal2017discrete}. 
\begin{equation}
    g_i(\b z_k, \b z_{k+1}) = h_i(\b z_{k+1}) - (1 - \alpha) h_i(\b z_k) - \delta
\end{equation}
The parameter $\alpha \in (0, 1)$ controls how quickly obstacles can be approached, and the parameter $\delta > 0$ robustifies the obstacle by a margin, enforcing a distance the plan must stay away from obstacle boundaries. Choosing $\alpha, \delta$ conservatively robustifies the plan against the tracking error which will occur when the plan is tracked on the full order system \cite{cohen2025safety}. 

\section{Experiment and Results}

We deploy the trained policy on hardware, and evaluate the impact of the environmentally consistent reference in simulation. Additionally, we demonstrate integration in a navigation autonomy stack by pairing the controller with an MPC controller through its $SE(2)$ velocity interface, performing indoor and outdoor navigation tasks autonomously. 

\subsection{Training and Deployment}
Policies are trained for 30k iterations over 4096 environments on a single NVIDIA 5080 GPU (16GB) in under 20 hours in the IsaacLab \cite{mittal2025isaaclab} training environment using the RSL-RL PPO implementation \cite{schwarke2025rslrl}. During deployment, depth reconstruction for the chest mounted ZED Mini runs on Unitree G1's integrated NVIDIA Jetson Orin. The policy is inferenced on CPU at 50Hz on an onboard System76 Meerkat minipc. When executing closed loop navigation control, we leverage a localization stack \cite{lio_localization} which combines FAST-LIO2 \cite{xu2021fast, xu2022fast} for odometry and lidar deskewing with GICP \cite{segal2009generalized} for localization within a pre-built map. These processes leverage the Livox MID360 LiDAR mounted in the head of the Unitree G1, and both run on the System76 Meerkat minipc. All sensing and compute for deployment is contained and powered onboard the robot. 

\begin{table}[t]
\centering
\caption{Cross-attention actor architecture. Depth imagery is fused with
proprioception via cross-attention with proprio-query and visual key/value.
Input depth image is $26{\times}30$.}
\label{tab:crossattn-arch}
\footnotesize
\setlength{\tabcolsep}{4pt}
\begin{tabular}{@{}p{0.18\columnwidth}p{0.18\columnwidth}p{0.38\columnwidth}p{0.16\columnwidth}@{}}
\toprule
\textbf{Module} & \textbf{Layer} & \textbf{Config} & \textbf{Output} \\
\midrule
Spatial CNN     & Conv2D $\times 2$    & ch.\ $[32,64]$, $5{\times}5$, stride $2$, zero-pad & $(64, 7, 8)$ \\
Token proj.     & Linear               & $64 \rightarrow 64$                                & $(56, 64)$ \\
Proprio enc.    & MLP, ELU             & $n_{\mathrm{prop}} \rightarrow 64 \rightarrow 64$  & $(64)$ \\
Cross-attn.     & MHA + LN             & $d{=}64$, $h{=}4$                                  & $(64)$ \\
Actor head      & MLP, ELU             & $128 \rightarrow 256 \rightarrow 128 \rightarrow n_a$ & $(n_a)$ \\
\bottomrule
\end{tabular}
\end{table}

\begin{table}[t]
\centering
\caption{Tracking error (norm of position error) by terrain.}
\label{tab:tracking}
\begin{tabular}{l ccc ccc}
\toprule
 & \multicolumn{3}{c}{Foot} & \multicolumn{3}{c}{CoM} \\
\cmidrule(lr){2-4} \cmidrule(lr){5-7}
Method & Flat & Slopes & Stairs & Flat & Slopes & Stairs \\
\midrule
unMod-Ref   & 0.037 & 0.055 & 0.142 & \textbf{0.020} & \textbf{0.027} & 0.083 \\
Ours        & 0.038 & 0.053 & \textbf{0.074} & 0.025 & 0.032 & \textbf{0.042} \\
Ours (Attn) & \textbf{0.032} & \textbf{0.045} & 0.074 & 0.021 & 0.032 & 0.045 \\
\bottomrule
\end{tabular}
\end{table}

\begin{figure*}
    \centering
    \includegraphics[width=1.0\linewidth]{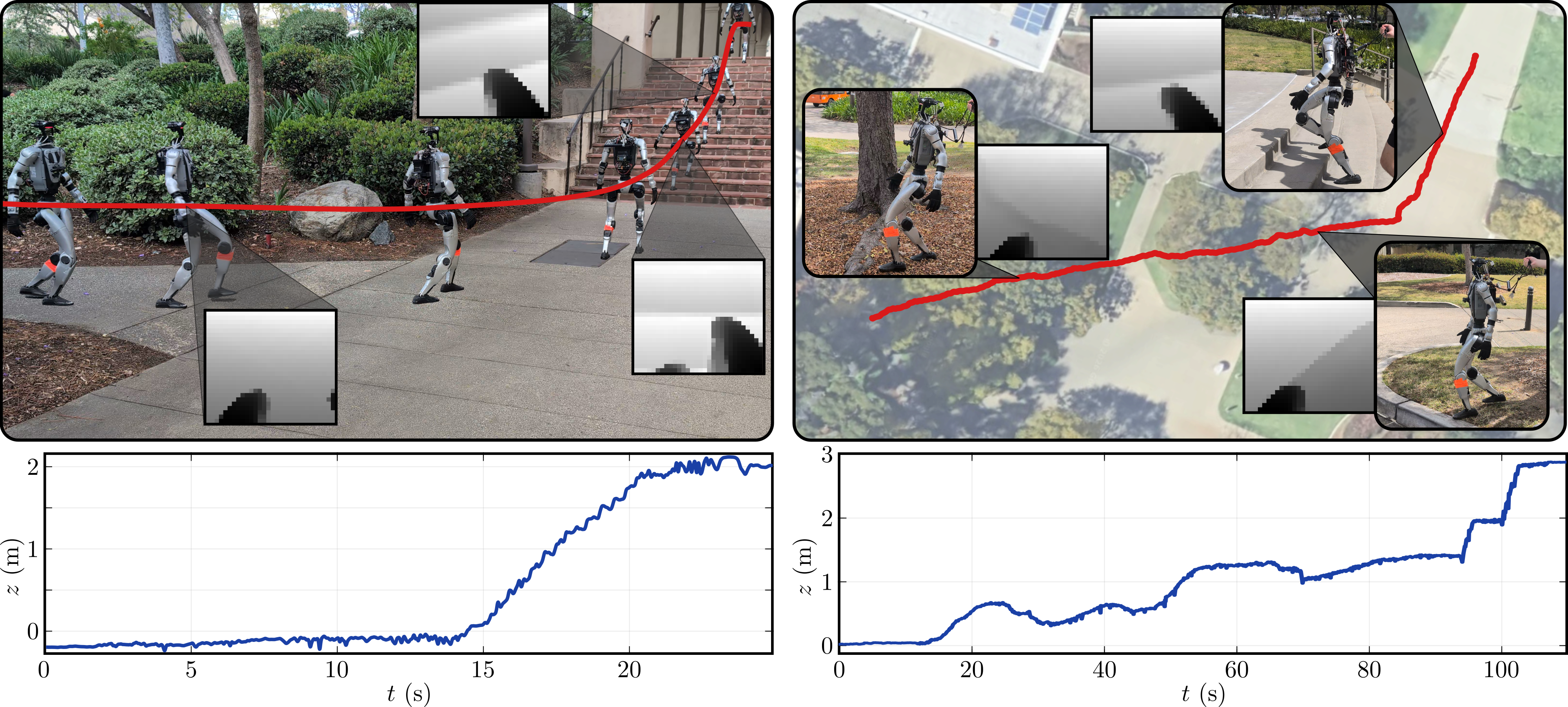}
    \caption{Outdoor autonomous navigation. The robot is localized within an existing map, while MPC plans trajectories through the environment over rough terrain and stairs. Depth images are rendered at a few frames, showing self-scans and terrain geometry. (Left) robot autonomously navigates 15 consecutive stairs. (Right) robot autonomously navigates over 70m, including two staircases.}
    \label{fig:hardware}
    \vspace{-5mm}
\end{figure*}

\subsection{Simulation Study}
We validated two key claims in simulation. First, conditioning the references on the environment yields significantly better tracking performance than tracking the nominal references alone. Second, unlike several reference-free papers in the current literature, we are able to achieve similar performance with a much simpler model. To complete this analysis we train two additional versions of our method; first we do not condition the reference on the environment (unMod-Ref). Second, we use an actor network with CNN and an attention-based perception encoder in an architecture similar to recent results in the field, whose architecture is described in \cref{tab:crossattn-arch}. The other policies use an MLP with layers $[1024, 512, 256, 128]$. Then, we compare the mean normed tracking error across flat, sloped, and stair terrains in \cref{tab:tracking}. Critically, Ours and Ours (Attn) have statistically similar tracking error in each environment type, but Ours without the attention head is much simpler to implement and inference, and requires significantly less VRAM during training (14Gb vs 60Gb). Additionally, we see that enforcing references to be consistent with the environment improves tracking error by a factor of two in the stairs environment. 

\subsection{Real-world deployment}

To emphasize the controllability of the learned locomotion policy, we integrate the policy with the MPC-based navigation planner detailed in \cref{sec:method}. First, a point cloud map of the relevant environment is created by scanning the environment with lidar while running FAST-LIO2. When the robot is deployed, FAST-LIO2 generates high-rate odometry, while GICP matches scans against the precomputed map at a low rate, compensating for potential odometry drift.

Obstacles in the environment are fit with ellipses or halfplane constraints to promote ease of transcription into a nonlinear program. 
The MPC problem is transcribed via CasADi \cite{Andersson2019} and solved online via IPOPT \cite{wachter2006implementation} once every two steps ($0.8s$). 
Inputs are constrained to lie in the set \eqref{eqn:ocp_input}
\begin{align*}
    \mathcal{V} = \{ \b v \in \R^3 ~|~ &v_\parallel \in [-0.3, 0.8],\\&v_\perp \in [-0.2, 0.2],~\omega \in [-0.8, 0.8]\}
\end{align*}
MPC outputs a discrete trajectory $(\b z_k, \b v_k)$. First, we construct a continuous $SE(2)$ trajectory ($\b z_{\text{MPC}}(t), \b v_{\text{MPC}}(t)$) obtained by linearly interpolating $\b z_k$ and sample-and-holding $\b v_k$. This continuous trajectory is then tracked via a feedback controller:
\begin{equation*}
    \b v(t) = \b R(t) \big(\underbrace{\b v_{\text{MPC}}(t)}_{ff} + \b K \underbrace{(\b z_{\text{MPC}}(t) - \b \Pi(\b x(t)))}_{error}\big)
\end{equation*}
This controller adds a feedback term on the system's tracking error to the feedforward velocity from the MPC plan, and rotates the linear components of the velocity into the robot's local frame through $\b R(t)$. We recompute this velocity command $\b v(t)$ at 10Hz, and pass it to the locomotion controller.

We successfully deploy this framework in several outdoor environments for long-horizon autonomous navigation. We autonomously navigate standard human staircases, including a section of 15 stairs (\cref{fig:hardware} left), and a trial spanning over 70m including tree roots, rough terrain, curbs, and two sets of stairs (\cref{fig:hardware} right). An additional trial, where the robot climbs a staircase, enters a building through a door, and then climbs indoor stairs, is pictured in \cref{fig:hero}.
As surveyed in \cref{tab:comparison}, prior humanoid navigation autonomy systems rely on blind, model-based locomotion, and prior perceptive locomotion methods have not been integrated into closed-loop autonomy at this scale. The deployment described here demonstrates a perception-conditioned, reference-guided RL policy serving as the locomotion layer of a long-horizon autonomy stack on a humanoid platform. 

\section{Conclusion}

We have presented a method for synthesizing environmentally-consistent reference trajectories inside the RL training loop for humanoid locomotion, and demonstrated its integration into a closed-loop navigation autonomy stack on the Unitree G1.
The proposed LIP-based reference synthesis is fast enough to run inside the training loop, and the resulting policy substantially reduces stair tracking error compared to tracking references which are not modified to be consistent with the environment.
Furthermore, the method exposes a useful $SE(2)$ velocity interface, enabling long-horizon outdoor deployments to demonstrate its efficacy as the locomotion layer of a humanoid autonomy stack in human-centric environments. 

Future work looks to extend this result in two directions.
First, modulating references from other sources, such as human data or trajectory optimization, to be consistent with arbitrary terrain online stands to reduce the demands of offline data collection, and improve the distribution of environments seen during training. 
Second, we plan to incorporate knowledge of the terrain-dependent capabilities of the locomotion controller (traversability, risk) into the planning layer to allow autonomous deployment of policies with highly varied and dynamic skill sets such as running, which add significant terrain-dependent constraints on what commands the mid-level controller can send to the locomotion policy.

\bibliographystyle{IEEEtran}
\balance
\bibliography{main.bib}

@IEEEtranBSTCTL{IEEEexample:BSTcontrol,
  CTLuse_forced_etal       = "yes",
  CTLmax_names_forced_etal = "6",
  CTLnames_show_etal       = "1"
}

@online{benGallantVoxelGridbased2025,
  title = {Gallant: {{Voxel Grid-based Humanoid Locomotion}} and {{Local-navigation}} across {{3D Constrained Terrains}}},
  shorttitle = {Gallant},
  author = {Ben, Qingwei and Xu, Botian and Li, Kailin and Jia, Feiyu and Zhang, Wentao and Wang, Jingping and Wang, Jingbo and Lin, Dahua and Pang, Jiangmiao},
  date = {2025-11-18},
  eprint = {2511.14625},
  eprinttype = {arXiv},
  eprintclass = {cs},
  doi = {10.48550/arXiv.2511.14625},
  url = {http://arxiv.org/abs/2511.14625},
  urldate = {2025-12-03},
  pubstate = {prepublished}
}

@online{wangBeamDojoLearningAgile2025,
  title = {{{BeamDojo}}: {{Learning Agile Humanoid Locomotion}} on {{Sparse Footholds}}},
  shorttitle = {{{BeamDojo}}},
  author = {Wang, Huayi and Wang, Zirui and Ren, Junli and Ben, Qingwei and Huang, Tao and Zhang, Weinan and Pang, Jiangmiao},
  date = {2025-04-27},
  eprint = {2502.10363},
  eprinttype = {arXiv},
  eprintclass = {cs},
  doi = {10.48550/arXiv.2502.10363},
  url = {http://arxiv.org/abs/2502.10363},
  urldate = {2025-12-03},
  pubstate = {prepublished}
}

@online{xiong3DUnderactuatedBipedal2021,
  title = {{{3D Underactuated Bipedal Walking}} via {{H-LIP}} Based {{Gait Synthesis}} and {{Stepping Stabilization}}},
  author = {Xiong, Xiaobin and Ames, Aaron},
  date = {2021-11-03},
  eprint = {2101.09588},
  eprinttype = {arXiv},
  eprintclass = {cs},
  doi = {10.48550/arXiv.2101.09588},
  url = {http://arxiv.org/abs/2101.09588},
  urldate = {2025-12-03},
  pubstate = {prepublished}
}

@online{xiongGlobalPositionControl2021,
  title = {Global {{Position Control}} on {{Underactuated Bipedal Robots}}: {{Step-to-step Dynamics Approximation}} for {{Step Planning}}},
  shorttitle = {Global {{Position Control}} on {{Underactuated Bipedal Robots}}},
  author = {Xiong, Xiaobin and Reher, Jenna and Ames, Aaron},
  date = {2021-11-29},
  eprint = {2011.06050},
  eprinttype = {arXiv},
  eprintclass = {cs},
  doi = {10.48550/arXiv.2011.06050},
  url = {http://arxiv.org/abs/2011.06050},
  urldate = {2025-12-03},
  pubstate = {prepublished}
}

@article{liao2025beyondmimic,
  title={Beyondmimic: From motion tracking to versatile humanoid control via guided diffusion},
  author={Liao, Qiayuan and Truong, Takara E and Huang, Xiaoyu and Gao, Yuman and Tevet, Guy and Sreenath, Koushil and Liu, C Karen},
  journal={arXiv preprint arXiv:2508.08241},
  year={2025}
}

@article{sleiman2026zest,
  title={ZEST: Zero-shot Embodied Skill Transfer for Athletic Robot Control},
  author={Sleiman, Jean Pierre and Li, He and Adu-Bredu, Alphonsus and Deits, Robin and Kumar, Arun and Bergamin, Kevin and Bhardwaj, Mohak and Biddlestone, Scott and Burger, Nicola and Estrada, Matthew A and others},
  journal={arXiv preprint arXiv:2602.00401},
  year={2026}
}

@article{allshire2025visual,
  title={Visual imitation enables contextual humanoid control},
  author={Allshire, Arthur and Choi, Hongsuk and Zhang, Junyi and McAllister, David and Zhang, Anthony and Kim, Chung Min and Darrell, Trevor and Abbeel, Pieter and Malik, Jitendra and Kanazawa, Angjoo},
  journal={arXiv preprint arXiv:2505.03729},
  year={2025}
}

@article{lee2024asap,
  title={ASAP: Agile and safe pursuit for local planning of autonomous mobile robots},
  author={Lee, Dong-Hyun and Choi, Sunglok and Na, Ki-In},
  journal={IEEe Access},
  volume={12},
  pages={99600--99613},
  year={2024},
  publisher={IEEE}
}

@article{li2026clf,
  title={Clf-rl: Control lyapunov function guided reinforcement learning},
  author={Li, Kejun and Olkin, Zachary and Yue, Yisong and Ames, Aaron D},
  journal={IEEE Robotics and Automation Letters},
  year={2026},
  publisher={IEEE}
}

@article{liu2025opt2skill,
  title={Opt2skill: Imitating dynamically-feasible whole-body trajectories for versatile humanoid loco-manipulation},
  author={Liu, Fukang and Gu, Zhaoyuan and Cai, Yilin and Zhou, Ziyi and Jung, Hyunyoung and Jang, Jaehwi and Zhao, Shijie and Ha, Sehoon and Chen, Yue and Xu, Danfei and others},
  journal={IEEE Robotics and Automation Letters},
  year={2025},
  publisher={IEEE}
}

@inproceedings{yu2022dynamic,
  title={Dynamic bipedal turning through sim-to-real reinforcement learning},
  author={Yu, Fangzhou and Batke, Ryan and Dao, Jeremy and Hurst, Jonathan and Green, Kevin and Fern, Alan},
  booktitle={2022 IEEE-RAS 21st International Conference on Humanoid Robots (Humanoids)},
  pages={903--910},
  year={2022},
  organization={IEEE}
}

@article{jenelten2024dtc,
  title={Dtc: Deep tracking control},
  author={Jenelten, Fabian and He, Junzhe and Farshidian, Farbod and Hutter, Marco},
  journal={Science Robotics},
  volume={9},
  number={86},
  pages={eadh5401},
  year={2024},
  publisher={American Association for the Advancement of Science}
}

@article{zhang2026rpl,
  title={RPL: Learning Robust Humanoid Perceptive Locomotion on Challenging Terrains},
  author={Zhang, Yuanhang and Seo, Younggyo and Chen, Juyue and Yuan, Yifu and Sreenath, Koushil and Abbeel, Pieter and Sferrazza, Carmelo and Liu, Karen and Duan, Rocky and Shi, Guanya},
  journal={arXiv preprint arXiv:2602.03002},
  year={2026}
}

@article{xu2021fast,
  title={Fast-lio: A fast, robust lidar-inertial odometry package by tightly-coupled iterated kalman filter},
  author={Xu, Wei and Zhang, Fu},
  journal={IEEE Robotics and Automation Letters},
  volume={6},
  number={2},
  pages={3317--3324},
  year={2021},
  publisher={IEEE}
}

@article{xu2022fast,
  title={Fast-lio2: Fast direct lidar-inertial odometry},
  author={Xu, Wei and Cai, Yixi and He, Dongjiao and Lin, Jiarong and Zhang, Fu},
  journal={IEEE Transactions on Robotics},
  volume={38},
  number={4},
  pages={2053--2073},
  year={2022},
  publisher={IEEE}
}

@inproceedings{segal2009generalized,
  author    = {A. Segal and D. Haehnel and S. Thrun},
  title     = {Generalized-{ICP}},
  booktitle = {Proceedings of Robotics: Science and Systems},
  year      = {2009},
  address   = {Seattle, USA},
  month     = {June},
  doi       = {10.15607/RSS.2009.V.021}
}

@misc{lio_localization,
  author       = {Compton, William},
  title        = {{LIO-Localization}: A {ROS2} workspace for {LiDAR}-Inertial Odometry-based mapping and localization},
  year         = {2026},
  howpublished = {\url{https://github.com/wdc3iii/LIO-Localization}},
  note         = {Accessed: 2026-04-26}
}

@article{schwarke2025rslrl,
  title={RSL-RL: A Learning Library for Robotics Research},
  author={Schwarke, Clemens and Mittal, Mayank and Rudin, Nikita and Hoeller, David and Hutter, Marco},
  journal={arXiv preprint arXiv:2509.10771},
  year={2025}
}

@article{mittal2025isaaclab,
  title={Isaac Lab: A GPU-Accelerated Simulation Framework for Multi-Modal Robot Learning},
  author={Mayank Mittal and Pascal Roth and James Tigue and Antoine Richard and Octi Zhang and Peter Du and Antonio Serrano-Muñoz and Xinjie Yao and René Zurbrügg and Nikita Rudin and Lukasz Wawrzyniak and Milad Rakhsha and Alain Denzler and Eric Heiden and Ales Borovicka and Ossama Ahmed and Iretiayo Akinola and Abrar Anwar and Mark T. Carlson and Ji Yuan Feng and Animesh Garg and Renato Gasoto and Lionel Gulich and Yijie Guo and M. Gussert and Alex Hansen and Mihir Kulkarni and Chenran Li and Wei Liu and Viktor Makoviychuk and Grzegorz Malczyk and Hammad Mazhar and Masoud Moghani and Adithyavairavan Murali and Michael Noseworthy and Alexander Poddubny and Nathan Ratliff and Welf Rehberg and Clemens Schwarke and Ritvik Singh and James Latham Smith and Bingjie Tang and Ruchik Thaker and Matthew Trepte and Karl Van Wyk and Fangzhou Yu and Alex Millane and Vikram Ramasamy and Remo Steiner and Sangeeta Subramanian and Clemens Volk and CY Chen and Neel Jawale and Ashwin Varghese Kuruttukulam and Michael A. Lin and Ajay Mandlekar and Karsten Patzwaldt and John Welsh and Huihua Zhao and Fatima Anes and Jean-Francois Lafleche and Nicolas Moënne-Loccoz and Soowan Park and Rob Stepinski and Dirk Van Gelder and Chris Amevor and Jan Carius and Jumyung Chang and Anka He Chen and Pablo de Heras Ciechomski and Gilles Daviet and Mohammad Mohajerani and Julia von Muralt and Viktor Reutskyy and Michael Sauter and Simon Schirm and Eric L. Shi and Pierre Terdiman and Kenny Vilella and Tobias Widmer and Gordon Yeoman and Tiffany Chen and Sergey Grizan and Cathy Li and Lotus Li and Connor Smith and Rafael Wiltz and Kostas Alexis and Yan Chang and David Chu and Linxi "Jim" Fan and Farbod Farshidian and Ankur Handa and Spencer Huang and Marco Hutter and Yashraj Narang and Soha Pouya and Shiwei Sheng and Yuke Zhu and Miles Macklin and Adam Moravanszky and Philipp Reist and Yunrong Guo and David Hoeller and Gavriel State},
  journal={arXiv preprint arXiv:2511.04831},
  year={2025},
  url={https://arxiv.org/abs/2511.04831}
}

@inproceedings{agrawal2017discrete,
  title={Discrete control barrier functions for safety-critical control of discrete systems with application to bipedal robot navigation.},
  author={Agrawal, Ayush and Sreenath, Koushil},
  booktitle={Robotics: Science and Systems},
  volume={13},
  pages={1--10},
  year={2017},
  organization={Cambridge, MA, USA}
}

@inproceedings{cohen2025safety,
  title={Safety-critical controller synthesis with reduced-order models},
  author={Cohen, Max H and Csomay-Shanklin, Noel and Compton, William D and Molnar, Tamas G and Ames, Aaron D},
  booktitle={2025 American Control Conference (ACC)},
  pages={5216--5221},
  year={2025},
  organization={IEEE}
}

@misc{stereolabs_zed_sdk,
  author       = {{Stereolabs}},
  title        = {{ZED} {SDK}},
  howpublished = {\url{https://www.stereolabs.com/developers/}},
  year         = {2024},
  note         = {Version 5.1}
}

@inproceedings{long2025learning,
  title={Learning humanoid locomotion with perceptive internal model},
  author={Long, Junfeng and Ren, Junli and Shi, Moji and Wang, Zirui and Huang, Tao and Luo, Ping and Pang, Jiangmiao},
  booktitle={2025 IEEE International Conference on Robotics and Automation (ICRA)},
  pages={9997--10003},
  year={2025},
  organization={IEEE}
}

@article{he2025attention,
  title={Attention-based map encoding for learning generalized legged locomotion},
  author={He, Junzhe and Zhang, Chong and Jenelten, Fabian and Grandia, Ruben and B{\"a}cher, Moritz and Hutter, Marco},
  journal={Science Robotics},
  volume={10},
  number={105},
  pages={eadv3604},
  year={2025},
  publisher={American Association for the Advancement of Science}
}

@article{zhuang2024humanoid,
  title={Humanoid parkour learning},
  author={Zhuang, Ziwen and Yao, Shenzhe and Zhao, Hang},
  journal={arXiv preprint arXiv:2406.10759},
  year={2024}
}

@article{wu2026perceptive,
  title={Perceptive humanoid parkour: Chaining dynamic human skills via motion matching},
  author={Wu, Zhen and Huang, Xiaoyu and Yang, Lujie and Zhang, Yuanhang and Sreenath, Koushil and Chen, Xi and Abbeel, Pieter and Duan, Rocky and Kanazawa, Angjoo and Sferrazza, Carmelo and others},
  journal={arXiv preprint arXiv:2602.15827},
  year={2026}
}

@inproceedings{su2025lipm,
  title={LIPM-Guided Reinforcement Learning for Stable and Perceptive Locomotion in Bipedal Robots},
  author={Su, Haokai and Luo, Haoxiang and Yang, Shunpeng and Jiang, Kaiwen and Zhang, Wei and Chen, Hua},
  booktitle={2025 IEEE-RAS 24th International Conference on Humanoid Robots (Humanoids)},
  pages={1031--1038},
  year={2025},
  organization={IEEE}
}

@article{agha2022nebula,
  title={NeBula: TEAM CoSTAR's robotic autonomy solution that won phase II of DARPA subterranean challenge},
  author={Agha, Ali and Otsu, Kyohei and Morrell, Benjamin and Fan, David D and Thakker, Rohan and Santamaria-Navarro, Angel and Kim, Sung-Kyun and Bouman, Amanda and Lei, Xianmei and Edlund, Jeffrey and others},
  journal={Field robotics},
  volume={2},
  pages={1432--1506},
  year={2022},
  publisher={FRPS}
}

@Article{Andersson2019,
  author = {Joel A E Andersson and Joris Gillis and Greg Horn
            and James B Rawlings and Moritz Diehl},
  title = {{CasADi} -- {A} software framework for nonlinear optimization
           and optimal control},
  journal = {Mathematical Programming Computation},
  volume = {11},
  number = {1},
  pages = {1--36},
  year = {2019},
  publisher = {Springer},
  doi = {10.1007/s12532-018-0139-4}
}

@article{wachter2006implementation,
  title={On the implementation of an interior-point filter line-search algorithm for large-scale nonlinear programming},
  author={W{\"a}chter, Andreas and Biegler, Lorenz T},
  journal={Mathematical programming},
  volume={106},
  number={1},
  pages={25--57},
  year={2006},
  publisher={Springer}
}

@article{yoon2025state,
  title={STATE-NAV: Stability-Aware Traversability Estimation for Bipedal Navigation on Rough Terrain},
  author={Yoon, Ziwon and Zhu, Lawrence Y and Lu, Jingxi and Gan, Lu and Zhao, Ye},
  journal={IEEE Robotics and Automation Letters},
  volume={11},
  number={2},
  pages={2338--2345},
  year={2025},
  publisher={IEEE}
}

@article{huang2023efficient,
  title={Efficient anytime clf reactive planning system for a bipedal robot on undulating terrain},
  author={Huang, Jiunn-Kai and Grizzle, Jessy W},
  journal={IEEE Transactions on Robotics},
  volume={39},
  number={3},
  pages={2093--2110},
  year={2023},
  publisher={IEEE}
}

@article{dixit2024step,
  title={Step: Stochastic traversability evaluation and planning for risk-aware navigation; results from the darpa subterranean challenge},
  author={Dixit, Anushri and Fan, David D and Otsu, Kyohei and Dey, Sharmita and Agha-Mohammadi, Ali-Akbar and Burdick, Joel},
  journal={Field Robotics},
  volume={4},
  pages={182--210},
  year={2024},
  publisher={FRPS}
}

@article{lin2021long,
  title={Long-horizon humanoid navigation planning using traversability estimates and previous experience},
  author={Lin, Yu-Chi and Berenson, Dmitry},
  journal={Autonomous Robots},
  volume={45},
  number={6},
  pages={937--956},
  year={2021},
  publisher={Springer}
}

@article{olkin2026chasing,
  title={Chasing Autonomy: Dynamic Retargeting and Control Guided RL for Performant and Controllable Humanoid Running},
  author={Olkin, Zachary and Compton, William D and Bena, Ryan M and Ames, Aaron D},
  journal={arXiv preprint arXiv:2603.25902},
  year={2026}
}

@article{cheng2024navila,
  title={Navila: Legged robot vision-language-action model for navigation},
  author={Cheng, An-Chieh and Ji, Yandong and Yang, Zhaojing and Gongye, Zaitian and Zou, Xueyan and Kautz, Jan and B{\i}y{\i}k, Erdem and Yin, Hongxu and Liu, Sifei and Wang, Xiaolong},
  journal={arXiv preprint arXiv:2412.04453},
  year={2024}
}

@inproceedings{lee2024integrating,
  title={Integrating model-based footstep planning with model-free reinforcement learning for dynamic legged locomotion},
  author={Lee, Ho Jae and Hong, Seungwoo and Kim, Sangbae},
  booktitle={2024 IEEE/RSJ International Conference on Intelligent Robots and Systems (IROS)},
  pages={11248--11255},
  year={2024},
  organization={IEEE}
}

\end{document}